\documentclass[10pt,twocolumn,letterpaper]{article}

\usepackage{cvpr}              % To produce the CAMERA-READY version
\definecolor{cvprblue}{rgb}{0.21,0.49,0.74}
\usepackage[pagebackref,breaklinks,colorlinks,allcolors=cvprblue]{hyperref}
\usepackage{pifont}
\usepackage{booktabs}
\usepackage{makecell}
\usepackage{multirow}
\usepackage{hyperref}
\usepackage[accsupp]{axessibility}  % Improves PDF readability for those with disabilities.
\usepackage[ruled,vlined]{algorithm2e}
\DontPrintSemicolon
\usepackage{amsmath}

\usepackage{fontawesome}

\def\paperID{34818} % *** Enter the Paper ID here
\def\confName{CVPR}
\def\confYear{2026}

\title{ResiHMR: Residual-Limb Aware Single-Image 3D Human Mesh Recovery\\ for Individuals with Limb Loss}

\author{{Jiaying Ying$^{\ast}$\textsuperscript{1}  \qquad Heming Du$^{\ast}$\textsuperscript{1} \qquad Kaihao Zhang\textsuperscript{2} \qquad Sean M. Tweedy \textsuperscript{1} \qquad Xin Yu$^{\dagger}$\textsuperscript{3} } \\
    \textsuperscript{1}The University of Queensland 
    \textsuperscript{2}Australian National University
    \textsuperscript{3}The University of Adelaide \\
    $^{\ast}$Equal contributors 
    $^{\dagger}$Corresponding author \\
    \tt\small uqjying@uq.edu.au \\
}

\begin{document}
\twocolumn[{%
\renewcommand\twocolumn[1][]{#1}%

\maketitle
\vspace{-30pt}

\begin{center}
    \centering
    \captionsetup{hypcap=false}
    \includegraphics[width=0.95\textwidth]{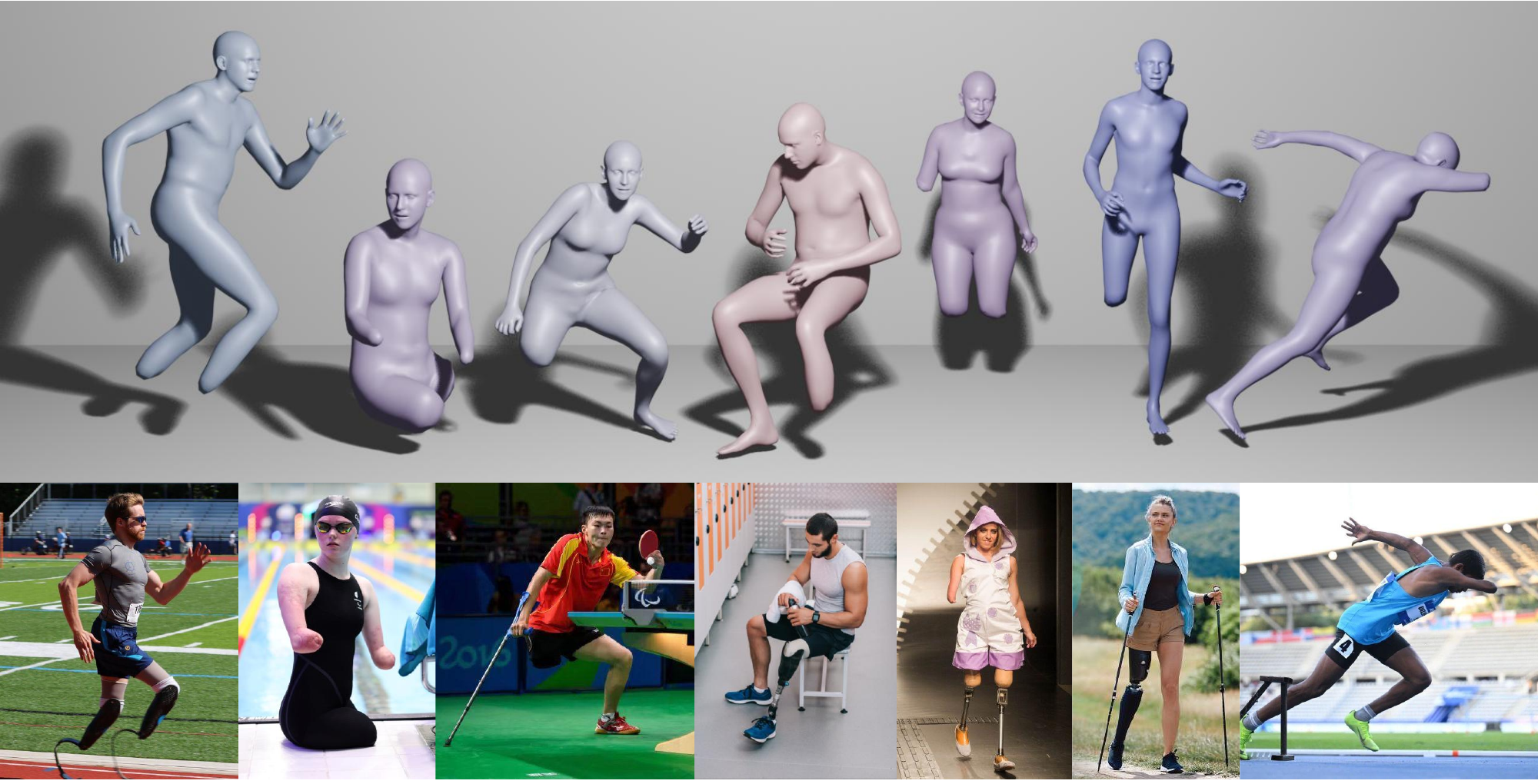}
    \captionof{figure}{
        \textbf{Demonstration of ResiHMR.} Our framework recovers anatomically coherent 3D body meshes from a single RGB image by adapting the kinematic topology and explicitly reconstructing residual-limb geometry.
    }
    \label{fig:title}
\end{center}
}]

\begin{abstract}

Single-image human mesh recovery provides a compact 3D, person-centric representation that supports analysis, animation, AR and VR, rehabilitation, and human–computer interaction. However, prevailing systems impose an intact-limb prior and degrade on people with limb loss, because fixed-topology models cannot represent residual limbs.
In this work, we present ResiHMR, a residual-limb aware framework for single-image 3D human modeling. ResiHMR adopts residual-limb keypoints and introduces two components: (i) a topology-adaptive Residual Anchor-Factor Optimization module that constrains estimation to the observed kinematic subgraph of anatomically valid structures, and (ii) a geometry-based Residual-Limb Reconstruction module that estimates residual-limb boundaries and convex limb-termination geometry. 
These components introduce topology-aware optimization and explicit termination geometry as tools for human mesh recovery under non-standard limb anatomy.
Unlike joint-removal methods in a fixed topology, ResiHMR explicitly reconstructs residual-limb surfaces and aligns optimization with limb-loss topology, which better matches prosthetic biomechanics and real-world use. 
To the best of our knowledge, this is the first single-image HMR system that explicitly reconstructs residual-limb surfaces and performs topology-adaptive optimization for individuals with limb loss. 
On a curated dataset of real-world images with limb loss, ResiHMR improves reconstruction quality under both SMPLify-X and HSMR backbones, reducing intact-joint 2D MPJPE from 41.32 to 37.40 with SMPLify-X and residual-limb 2D MPJPE from 73.61 to 23.19 with HSMR.
\\\href{https://akitaraphael.github.io/ResiHMR/}{\faGithub~\textcolor{purple}{ResiHMR}}

\end{abstract}    
\section{Introduction}
\label{sec:intro}

\begin{figure}[t]
    \centering
    \includegraphics[width=1\linewidth]{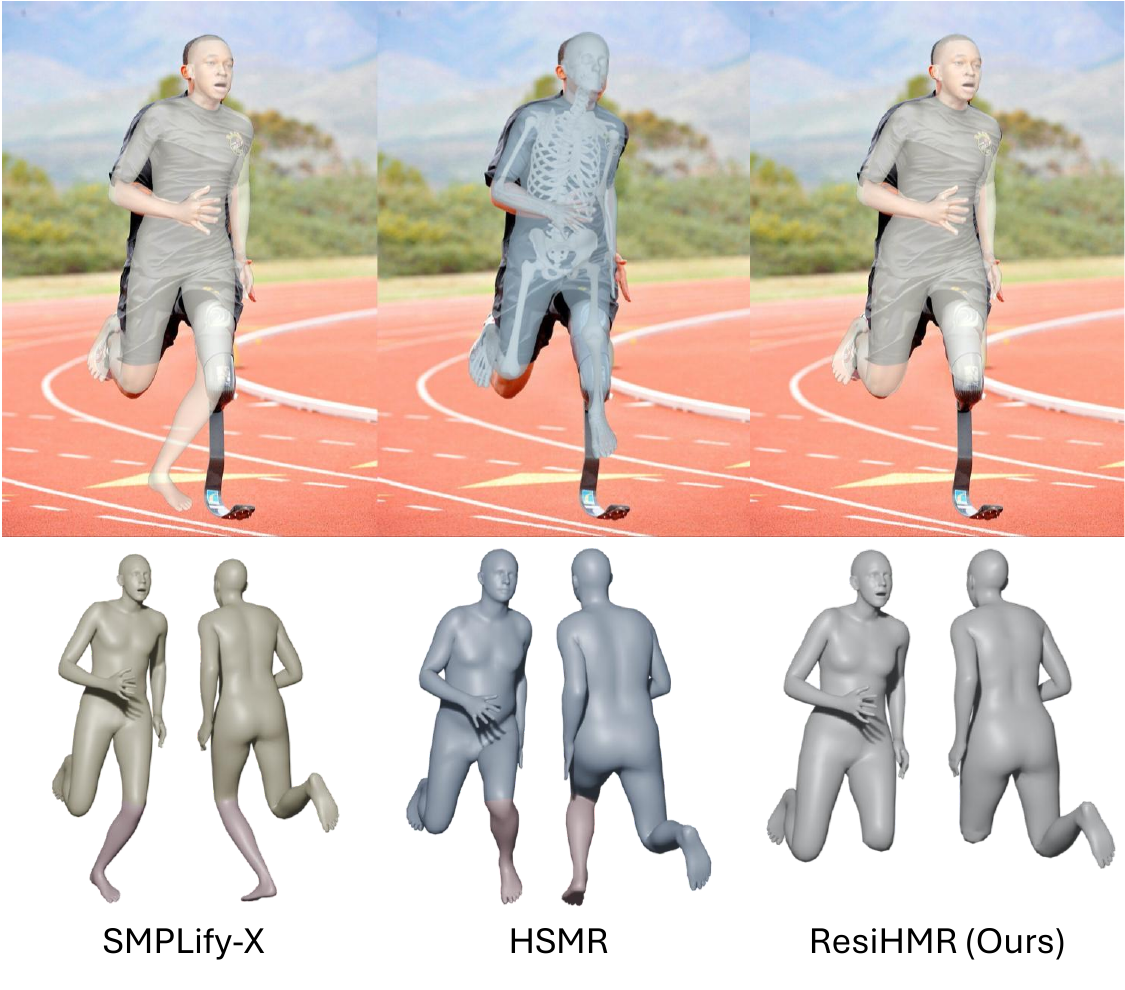}
    \caption{\textbf{Failure cases of existing HMR methods on individuals with limb loss.}
    SMPLify-X and HSMR hallucinate intact limbs or distort the lower body due to their intact-limb priors and fixed kinematic topologies. In contrast, ResiHMR correctly localizes the residual-limb, avoids compensatory distortions, and reconstructs an anatomically coherent body mesh.}
    \label{fig:failedcase}
\end{figure}

Single image human mesh recovery provides a compact, person centric 3D representation that supports animation and free viewpoint rendering~\citep{weng2022humannerf,zhu2024champ}, sports analysis~\citep{jiang2024worldpose}, and rehabilitation or clinical monitoring~\citep{smpla}. Parametric body models such as SMPL~\citep{smpl2023} and SMPL X~\citep{smplifyx2019} make this practical by providing structured pose and shape spaces for optimization and learning, while HSMR~\citep{hsmr} and SKEL~\citep{skel2023} target biomechanical interpretability by regressing a skeleton designed for physics based simulation.

Despite these benefits, existing HMR pipelines inherit an intact-limb assumption from fixed-topology body models and their priors. The skeletal graph, vertex connectivity, and pose priors are defined for able-bodied anatomy, so reconstruction degrades on people with limb loss and cannot represent residual-limb surfaces or anatomically meaningful limb terminations. This causes hallucinated segments and unstable optimization around the amputation site. The affected population is large, with global studies reporting tens of millions of people living with traumatic limb amputation and an increasing burden over time~\citep{McDonald2021GlobalAmputation,Wei2025TraumaticAmputationBurden,Rivera2024USLimbLossPrevalence}. These facts motivate methods that reconstruct residual-limb geometry instead of masking or removing joints and that provide 3D representations suitable for prosthetic alignment, socket assessment, gait analysis, and rehabilitation monitoring.

As Figure~\ref{fig:failedcase} shows, methods that assume intact anatomy mis-handle individuals with limb loss. On a left transfemoral example, SMPL-X-based fits hallucinate a full thigh and shank. HSMR regresses a full-body biomechanical skeleton, so errors around the residual-limb disturb the skeleton and further degrade the reconstruction of intact limbs.
To tackle this challenge, we propose ResiHMR, a residual-limb aware single-image framework that reconstructs anatomically coherent meshes for people with limb loss. To the best of our knowledge, ResiHMR is the first system that explicitly models residual-limb surfaces and adapts optimization to limb-loss topology, rather than preserving a fixed full-limb graph. This design targets common cases in prosthetics and rehabilitation in which the quality of the limb-termination geometry matters.

ResiHMR uses residual-limb keypoints~\citep{ldpose} within HMR and introduces two components. Residual Anchor-Factor Optimization component constrains estimation to anatomically valid kinematic subgraphs, which reduces spurious joint regression on absent limbs and stabilizes optimization near the amputation site. 
Residual-Limb Reconstruction component predicts residual-limb boundaries and smooth, convex stump surfaces, giving interpretable residual-limb length and physically plausible termination geometry. 
Because the residuum interacts with the socket to bear load and transmit control, and this interface is central to comfort, fit, and function in clinical biomechanics, explicit stump geometry allows ResiHMR to better match prosthetic biomechanics and to provide residual-limb aware 3D representations that support prosthetists, rehabilitation clinicians, and researchers in real workflows.

We evaluate ResiHMR on a curated LDPose-LimbLoss Evaluation Dataset with 2D residual-limb and intact keypoints and pixel masks. We compare against the recent HMR methods. We report 2D reprojected mesh error, 2D joint error on intact joints, 2D residual-endpoint localization error, and silhouette consistency in the form of mask IoU. ResiHMR reduces 2D mesh error, improves residual-endpoint alignment, and yields more anatomically coherent stump geometry in qualitative results.
This work makes the following contributions:
\begin{itemize}
    \item We propose ResiHMR, a single image residual-limb aware HMR framework that reconstructs anatomically coherent 3D human meshes for people with limb loss by explicitly modeling residual-limb surfaces and adapting optimization to limb loss topology.

    \item We introduce a Residual Anchor-Factor Optimization component and a Residual-Limb Reconstruction component that estimate residual-limb boundaries and convex termination geometry.

    \item We provide a topology adaptive residual-limb aware 3D body representation that better matches prosthetic biomechanics and supports prosthetic alignment, rehabilitation, and mobility analysis on a curated LDPose-LimbLoss Evaluation Dataset.
\end{itemize}

\section{Related Work}
\label{sec:rw}

\begin{figure*}[t]
    \centering
    \includegraphics[width=\linewidth]{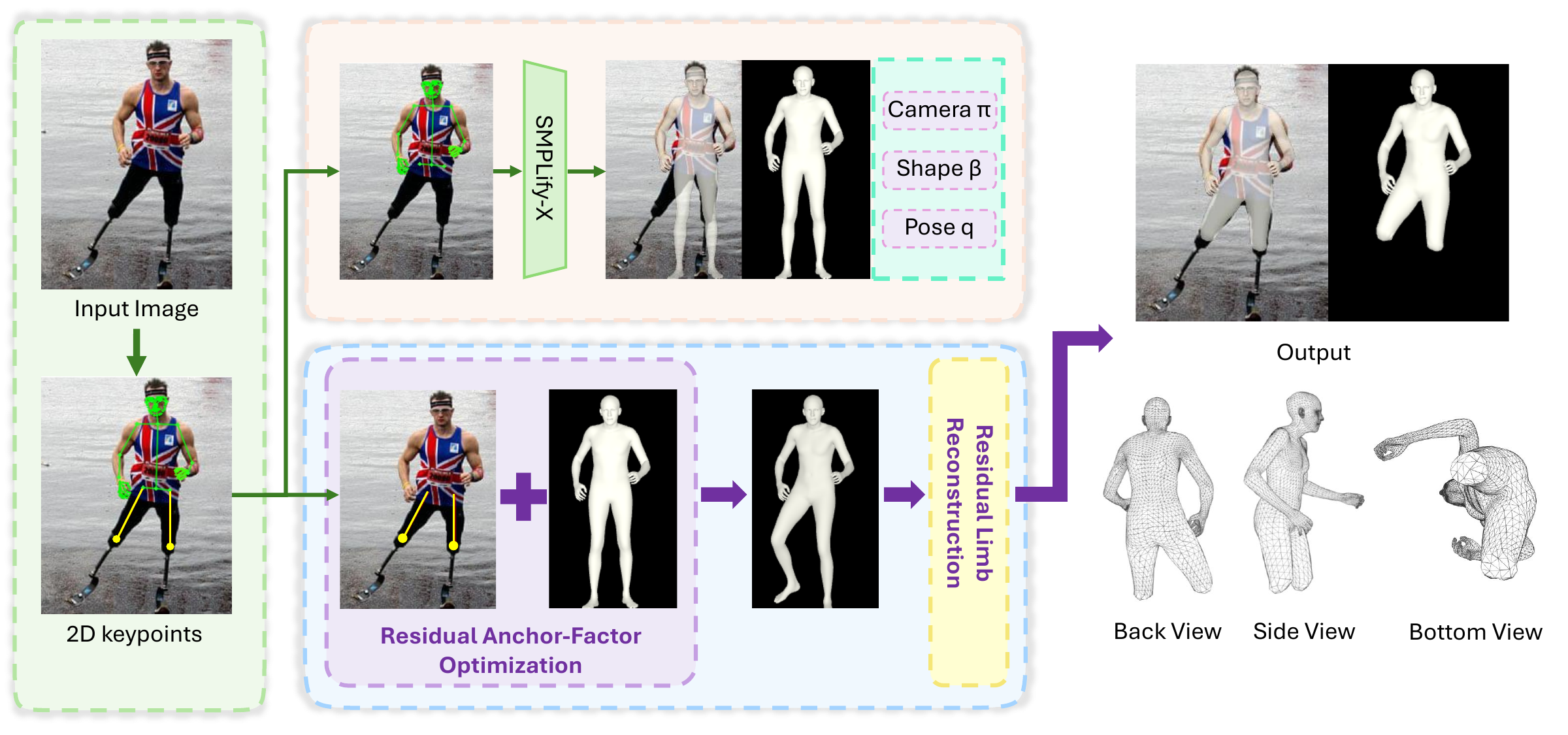}
    \vspace{-2em}
    \caption{\textbf{Overview of our ResiHMR framework.}Given an input image, SMPL-X is initialized using intact 2D keypoints. Our Residual Anchor–Factor Optimization adapts the kinematic graph by refining anchor joints and residual-limb proportions under supervision of residual-limb 2D keypoints. The Residual-Limb Reconstruction module then removes distal limb geometry and generates a smooth, watertight stump surface, producing anatomically realistic residual-limb aware meshes. }
    \label{fig:pipeline}
    \vspace{-1em}
\end{figure*}

\subsection{Human Body Models}

Parametric human body models provide a compact, controllable representation of shape and articulation. Early models such as SCAPE~\cite{scape} learned pose-dependent deformations from 3D scans, while SMPL~\cite{smpl2023} introduced a differentiable, low-dimensional body model that has become central to vision and graphics. Subsequent extensions model specific regions, including MANO for hands~\cite{MANO}, FLAME for face~\cite{FLAME}, and SMPL-X~\cite{smplifyx2019} unifies body, hands, and face within a single expressive mesh. Recent work improves realism and population coverage: SKEL~\cite{skel2023} incorporates a biomechanically grounded skeleton, SMPLer-X~\cite{smpler} scales whole-body modeling to large datasets, STAR~\cite{star} enhances pose-dependent deformation, GHUM~\cite{GHUM} introduces a generative articulated prior, and Anny~\cite{anny} expands demographic diversity.

While appearance driven reconstruction approaches such as NeRFs \cite{nerf,su2021nerf,hu2023sherf}, implicit SDF models \cite{park2019deepsdf,cao2022jiff}, and 4D Gaussian splatting \cite{wu20244d} produce highly realistic geometry but lack anatomical structure such as joints, kinematic chains, or interpretable limb measurements. Hybrid methods such as ICON \cite{icon}, SiTH \cite{sith}, and ECON \cite{econ} preserve SMPL controllability while refining surface detail, but they still assume intact-limb topology. 

However, despite covering broader shape variation and more realistic articulation, all existing parametric body models assume intact limb topology in both mesh representation and kinematic chain structure, making them unable to represent residual-limb geometry.

\subsection{Single-Image Human Mesh Recovery}

Single-image 3D human mesh recovery has advanced rapidly in recent years.
Classical optimization-based methods such as SMPLify~\cite{smplify} and SMPLify-X~\cite{smplifyx2019} fit a parametric model to 2D keypoints using strong pose and shape priors, but inference is slow and sensitive to keypoint noise.
Regression-based approaches predict model parameters directly from images, beginning with HMR~\cite{HMR18} and HMR2.0~\cite{HMR20}, and later refined through in-the-loop optimization (SPIN~\cite{spin}), occlusion reasoning (PARE~\cite{pare}), improved camera modeling (CLIFF~\cite{cliff}), and iterative alignment modules (PyMAF-X~\cite{pymafx2023}, ProHMR~\cite{prohmr}).
Hybrid analytical–neural methods such as HybrIK~\cite{hybrik} and HybrIK-X~\cite{hybrikx} integrate differentiable inverse kinematics for more accurate rotations, while transformer-based whole-body models (MeshGraphormer~\cite{lin2021-mesh-graphormer}, METRO~\cite{lin2021end}, SMPLer-X~\cite{smpler}, OSX~\cite{osx}) further push reconstruction fidelity.
PHYS-HMR~\cite{physhmr} augments visual cues with physics-based humanoid control to improve motion realism. Despite these advances, all methods assume an intact-limb topology inherited from SMPL~\cite{smpl2023}/SMPL-X~\cite{smplifyx2019}.

\paragraph{3D Human Modeling for Individual with Disabilities}

The applicability of 3D human body models to individuals with limb loss remains underexplored. Clinical studies show that residual-limb geometry is essential for prosthetic design, rehabilitation, and gait assessment~\cite{canda2009stature,connick2016evaluation}, highlighting the need for anatomically grounded models.

Recently, AJAHR\footnote{At the time of submission, the official implementation of AJAHR was not publicly released, preventing a fair and reproducible comparison. We therefore do not include AJAHR in our experimental evaluation. Additional discussion appears in the Supplementary Material.} \cite{ajahr} introduces joint-existence prediction for amputee-aware SMPL fitting. However, it still relies on intact SMPL topology, yielding stump locations at intact joints due to missing residual-limb endpoints.
The associated A3D and ITW-Amputee datasets annotate joint presence but lack anatomical residual-limb endpoints, leading reconstructed limbs to terminate at the nearest intact SMPL joint rather than at the observed stump. 
LDPose~\cite{ldpose} complements this direction by introducing a 2D keypoint system for individuals with limb loss, defining residual limb endpoints at the image level. However, it does not include 3D supervision or a parametric body representation.

As a result, no existing method or dataset supports anatomically grounded 3D reconstruction for individuals with limb loss, where residual-limbs are explicitly modeled rather than implicitly shortened. Our work provides the first residual-limb–aware extension of parametric single-image 3D reconstruction.

\section{ResiHMR Framework}
\label{sec:method}

We propose \textbf{ResiHMR}, a residual-limb aware 3D human mesh recovery framework that estimates anatomically coherent meshes from a single RGB image. ResiHMR is a fully optimization-based framework and does not require any training dataset. Given an image and whole-body 2D keypoints in the LDPose format, ResiHMR first fits a SMPL-X model~\cite{smplifyx2019} to intact keypoints, then applies a \textbf{Residual Anchor-Factor Optimization} module to adapt the kinematic graph and a geometry-based \textbf{Residual-Limb Reconstruction} module to remove distal limb geometry and generate a smooth, convex, watertight stump surface, as illustrated in Figure~\ref{fig:pipeline}.

\subsection{Preliminaries}

\textbf{SMPL-X and Parameterized Backbones.}
We build upon SMPL-X~\cite{smplifyx2019}, a differentiable parametric model of the body, hands, and face that represents humans as a deformable mesh driven by a kinematic skeleton. As our primary optimization backbone, we adopt SMPLify-X due to its explicit and interpretable formulation, which optimizes camera, pose, and shape parameters through modular objectives and accommodates additional supervision and constraints. However, SMPL-X assumes fixed intact-limb topology with full-limb kinematic chains. When directly fitted to residual-limb keypoints, it tends to hallucinate missing distal segments and place residual endpoints at the nearest intact joint, leading to anatomically incorrect meshes and unstable optimization near the limb termination.

More importantly, our framework is not restricted to SMPLify-X. 
ResiHMR is compatible with any HMR pipeline that outputs SMPL-X parameters (camera, pose, and shape), including both optimization-based and regression-based methods.
Our formulation operates directly on the parameter space and kinematic structure, enabling flexible incorporation of residual-limb keypoints, anchor redefinition, and amputation-aware constraints.
This design makes ResiHMR a plug-in module for existing HMR systems, extending them to support anatomically coherent reconstruction for individuals with limb loss.

\noindent\textbf{LDPose.}
To represent individuals with limb loss, we use the LDPose~\cite{ldpose} keypoint schema\footnote{For more details, please refer to Figure 2 of LDPose}, which augments intact 2D body keypoints with eight residual-limb endpoints. LDPose provides 2D keypoints, so reconstructing a 3D residual limb requires adapting SMPL-X kinematics to limb absence.

\subsection{Problem Definition}

Our goal is to lift 2D residual-limb endpoints into anatomically meaningful 3D limb terminations while keeping global pose and body proportions consistent with intact keypoints. The reconstructed mesh should respect the SMPL-X skeletal hierarchy and give each residual limb an explicit 3D geometry instead of implicitly shortening or omitting it.

Given a single RGB image $I \in \mathbb{R}^{H \times W \times 3}$ and detected 2D keypoints
$\mathbf{K}_{2D} = \{(x_i, y_i, c_i)\}_{i=1}^{N+M}$,
we reconstruct a 3D human mesh that represents both intact parts and residual-limbs.
Each keypoint $(x_i, y_i)$ is a pixel coordinate with confidence $c_i$.
We use OpenPose~\cite{openpose} WholeBody keypoints
$\mathbf{K}_{2D}^{\text{intact}} = \{(x_j, y_j, c_j)\}_{j=1}^{N}$ with $N = 135$ body, hand, and face keypoints
and extend them with residual-limb keypoints
$\mathbf{K}_{2D}^{\text{residual}} = \{(x_j, y_j, c_j)\}_{j=1}^{M}$ with $M = 8$ following LDPose~\cite{ldpose}.
We estimate the parameters of a SMPL-X model $\Theta = \{\boldsymbol{\theta}, \boldsymbol{\beta}, \boldsymbol{\psi}, \mathbf{R}, \mathbf{t}, \lambda_r\}$,
where $\boldsymbol{\theta}$, $\boldsymbol{\beta}$, and $\boldsymbol{\psi}$ are pose, shape, and expression,
$\mathbf{R} \in \mathrm{SO}(3)$ and $\mathbf{t} \in \mathbb{R}^3$ are global rotation and translation,
and $\lambda_r \in [0,1]$ encodes a proportional residual-limb length factor along the kinematic chain.
The resulting mesh $\mathbf{M}_r(\Theta)$ should combine intact and residual geometry in an anatomically coherent way.

\subsection{Residual Anchor-Factor Optimization}

We build on SMPLify-X~\cite{smplifyx2019} as our primary optimization backbone. 
The optimization-based formulation adapts more easily to unseen body topologies such as limb loss and it keeps full access to SMPL-X pose and shape parameters, which we later use for explicit geometry editing. 
In the experiments and supplementary material, we further demonstrate that ResiHMR can also be built on alternative HMR backbones, such as the regression-based HSMR model.

We first use intact keypoints $\mathbf{K}_{2D}^{\text{intact}}$ to initialize SMPL-X camera, pose, and shape following the standard SMPLify-X objective:
\begin{equation}
\min_{\boldsymbol{\theta}, \boldsymbol{\beta}, \mathbf{t}} \; E_{\text{data}} + E_{\text{prior}},
\end{equation}
where the data term matches projected joints to detected keypoints:
\begin{equation}
E_{\text{data}} = \sum_{i} w_i \left\| \pi(\mathbf{J}_i) - \mathbf{k}_i^{2D} \right\|^2,
\end{equation}
and $E_{\text{prior}}$ regularizes pose and shape with learned priors. Here $\mathbf{J}_i$ are 3D joints and $\mathbf{k}_i^{2D}$ are detected intact keypoints.

Residual keypoints $\mathbf{K}_{2D}^{\text{residual}}$ and the SMPLify-X initialization $(\boldsymbol{\theta}, \boldsymbol{\beta}, \boldsymbol{\pi})$ define the Residual Anchor Factor Optimization stage. For each residual-limb we jointly optimize an \emph{anchor joint} position $\mathbf{J}_a$ and a \emph{residual length factor} $\lambda_r \in [0,1]$. The anchor joint is the distal joint at the base of the amputated limb, for example the knee or elbow. The factor $\lambda_r$ specifies where the limb terminates on the line from $\mathbf{J}_a$ to its upstream joint $\mathbf{J}_t$. We follow anthropometric findings~\cite{canda2009stature,connick2016evaluation} that intact segment lengths correlate with residual-limb length and we therefore keep the intact limb length from SMPLify-X as a prior while only adjusting the termination point.

\noindent\textbf{Residual representation.}
We represent each residual-limb $r$ by an endpoint that lies on the kinematic line segment between the proximal anchor joint and its upstream joint. Concretely, we define
\begin{equation}
\mathbf{R}_r = \mathbf{J}_a + \lambda_r \, (\mathbf{J}_t - \mathbf{J}_a),
\end{equation}
where $\mathbf{J}_a$ is the anchor joint, $\mathbf{J}_t$ is its upstream joint, and $\lambda_r \in [0,1]$ encodes the remaining limb proportion. A value $\lambda_r = 0$ corresponds to an intact limb and $\lambda_r = 1$ indicates complete removal. This linear interpolation yields a low-dimensional and anatomically grounded parameterization of the residual-limb. It ties the endpoint to the underlying kinematic chain, preserves the direction of the original limb segment, and lets the optimizer adjust residual-limb length through a single factor while keeping the global skeleton structure consistent with SMPL-X.

\noindent\textbf{Loss function and optimization.}
For each visible residual keypoint we optimize $(\mathbf{J}_a,\lambda_r)$ with
\begin{equation}
\mathcal{L} = \mathcal{L}_{\text{reproj}} + \alpha\,\mathcal{L}_{\text{reg}} + \mu\,\mathcal{L}_{\text{len}},
\end{equation}
where $\alpha$ and $\mu $ are adaptively scaled based on the initial fitting error for each instances. The reprojection loss enforces 2D consistency between the residual endpoint and its observation, with $\mathcal{L}_{\text{reproj}} = \|\pi(\mathbf{R}_r) - \mathbf{k}_r^{2D}\|^2$. The regularizer penalizes large anchor motion away from the SMPLify-X estimate, $\mathcal{L}_{\text{reg}} = \|\mathbf{J}_a - \mathbf{J}_a^{\text{init}}\|^2$, so that intact joints remain close to the original kinematic structure. The length term preserves the limb length given by SMPLify-X and encodes anthropometric priors,
\begin{equation}
\mathcal{L}_{\text{len}} =
\left(\left\| \mathbf{J}_t - \mathbf{J}_a \right\|
- \left\| \mathbf{J}_t - \mathbf{J}_a^{\text{init}} \right\|\right)^2,
\end{equation}
which encourages biomechanically reasonable residual segments.

We use L-BFGS~\cite{liu1989limited} with strong Wolfe line search. We optimize each residual-limb independently and accept the solution if the residual endpoint reprojection error is below a threshold, which is 15 pixels in practice. The procedure of Residual Anchor--Factor Optimization for a single residual-limb is summarized in Algorithm~\ref{alg:anchor_factor}.

\begin{algorithm}[t]
\caption{Residual Anchor--Factor Optimization}
\label{alg:anchor_factor}
\DontPrintSemicolon
\KwIn{Initial anchor joint $\mathbf{J}_a^{\text{init}}$, target joint $\mathbf{J}_t$, residual 2D keypoint $\mathbf{k}_{2D}$, camera $\pi(\cdot)$}
\KwOut{Optimized anchor $\mathbf{J}_a^{*}$ and residual factor $\lambda^{*}$}

Initialize $\mathbf{J}_a \leftarrow \mathbf{J}_a^{\text{init}}$, $\lambda \leftarrow 0.5$ \;
\For{$i = 1$ \KwTo $N$ (L-BFGS iterations)}{
    % Compute residual limb point in 3D
    $\lambda \leftarrow \mathrm{clip}(\lambda, \lambda_{\min}, \lambda_{\max})$\;
    $\mathbf{R} = \mathbf{J}_a + \lambda \cdot (\mathbf{J}_t - \mathbf{J}_a)$\;
    
    % Project to 2D
    $\mathbf{r}_{2D} = \pi(\mathbf{R})$\;
    
    % Loss terms
    $\mathcal{L}_{\text{reproj}} = \|\mathbf{r}_{2D} - \mathbf{k}_{2D}\|^2$\;
    $\mathcal{L}_{\text{reg}} = \|\mathbf{J}_a - \mathbf{J}_a^{\text{init}}\|^2$\;
    $\mathcal{L}_{\text{len}} = \left(\|\mathbf{J}_t - \mathbf{J}_a\| - \|\mathbf{J}_t - \mathbf{J}_a^{\text{init}}\|\right)^2$\;
    
    % Total objective
    $\mathcal{L} = \mathcal{L}_{\text{reproj}} + \alpha \mathcal{L}_{\text{reg}} + \mu \mathcal{L}_{\text{len}}$\;
    
    % Update (L-BFGS)
    $(\mathbf{J}_a, \lambda) \leftarrow \text{L-BFGS-step}(\mathbf{J}_a, \lambda, \nabla \mathcal{L})$\;
}
\Return $\mathbf{J}_a^{*} = \mathbf{J}_a,\ \lambda^{*} = \lambda$ \;
\end{algorithm}

\subsection{Residual-Limb Reconstruction}

Given the optimized anchor joints and residual length factors $(\mathbf{J}_a^\star,\lambda_r^\star)$ from the Residual Anchor Factor Optimization stage and the SMPL-X mesh $\mathbf{M} = (\mathbf{V}, \mathbf{F})$, we modify the mesh to remove distal limb geometry and construct an explicit stump surface. For each residual-limb $r$ we first compute a 3D cut point: 
\begin{equation}
\mathbf{p}_r = \mathbf{J}_a^\star + \lambda_r^\star \bigl(\mathbf{J}_t^{\text{init}} - \mathbf{J}_a^\star\bigr),
\end{equation}
where $\mathbf{J}_t^{\text{init}}$ is the upstream joint in the original SMPL-X skeleton. All subsequent operations are restricted to the corresponding limb submesh.

\begin{figure*}[t]
    \centering
    \includegraphics[width=1\linewidth]{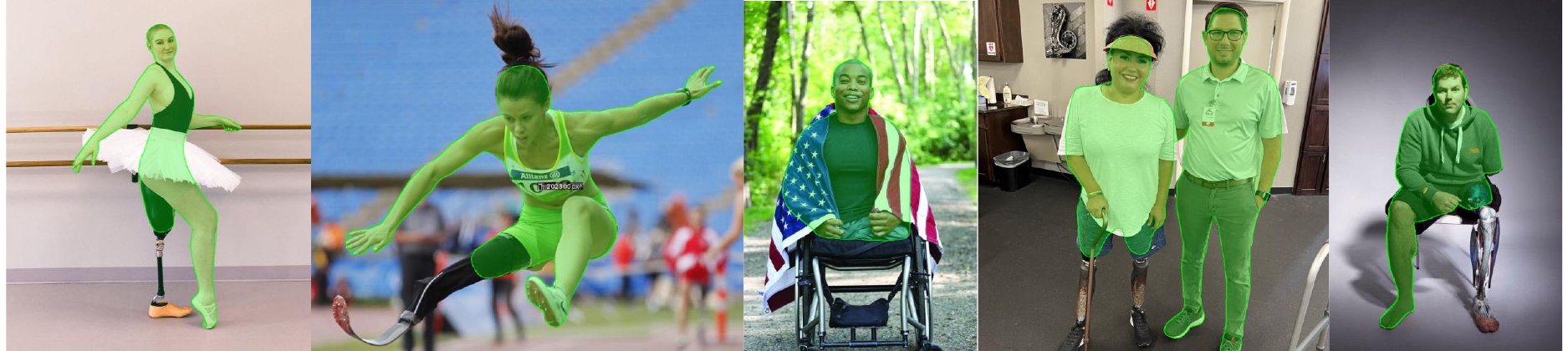}
    \caption{\textbf{Dataset Demonstration of the proposed LDPose-LimbLoss Evaluation Dataset.} Representative samples illustrating the diversity of subjects, amputation levels, poses, activities, and environments included in the dataset. \textcolor{green}{Green} overlays indicate the manually annotated body masks used to isolate the human body region.}
    \label{fig:dataset}
    % \vspace{-1.5em}
\end{figure*}

\noindent\textbf{Segmentation guided coarse pruning.}
We apply a SMPL-X body parts segmentation and remove distal part labels for the residual-limb, for example forearm and hand or foot, to obtain a coarse distal mask. This step discards vertices that clearly lie beyond the intended residual segment and reduces the search space for the subsequent geometric cut.

\noindent\textbf{Fine geometric cutting.}
Within the preserved segment region, for example upper arm or thigh, we identify the face nearest to $\mathbf{p}_r$ and grow a connected vertex ring inside a narrow band around a cut plane. We define the plane normal:
\begin{equation}
\hat{\mathbf{n}} = \frac{\mathbf{J}_a^\star - \mathbf{J}_t^{\text{init}}}{\left\|\mathbf{J}_a^\star - \mathbf{J}_t^{\text{init}}\right\|}.
\end{equation}
For each vertex $\mathbf{v}$ in the kept region we compute the signed distance
$\phi(\mathbf{v}) = \langle \mathbf{v} - \mathbf{p}_r, \hat{\mathbf{n}} \rangle$.
Vertices with $\phi(\mathbf{v})$ above a small margin that are not part of the protective ring are treated as distal geometry and removed. This procedure yields a smooth intersection contour and is robust to local surface irregularities and mesh noise.

\noindent\textbf{Boundary cleanup and sealing.}
We extract boundary vertices, that is edges with multiplicity one, and iteratively prune low degree boundary vertices to remove spurious spikes. We then fit a local plane and generate two concentric vertex rings with small offsets $\pm h$ along the plane normal. Triangulation between the boundary and these rings produces a smooth, convex, watertight stump surface. The resulting mesh $\mathbf{M}_r(\Theta)$ remains closed and is visually consistent with limb termination anatomy.

This reconstruction stage uses the residual parameters inferred in the previous optimization stage to drive explicit mesh editing, so the final meshes have anatomically meaningful stump geometry that is suitable for clinical assessment, prosthetic alignment, and downstream analysis.

\subsection{LDPose-LimbLoss Evaluation Dataset}

To evaluate ResiHMR and support future work on single image mesh recovery for people with limb loss, we construct the LDPose-LimbLoss Evaluation Dataset, a curated subset of LDPose~\cite{ldpose} with extended annotations. The dataset contains 255 images that cover diverse upper and lower limb loss configurations, amputation levels, residual-limb lengths, activities, and backgrounds with limited self occlusion. Each image includes 2D annotations of 17 standard body keypoints and 8 residual-limb endpoints, and we add per person segmentation masks to isolate the human region and reduce the influence of clothing, hair, prosthetic devices, and other objects. For partially occluded subjects we estimate missing body regions from visible context to keep anatomical completeness, and for each residual-limb we annotate the termination segment based on LDPose residual keypoints and visual inspection. Figure~\ref{fig:dataset} illustrates examples, and full dataset statistics appear in the Supplementary Material.

\section{Experiments}
\label{sec:experiments}

\begin{figure*}[t]
    \centering
    \includegraphics[width=0.95\linewidth]{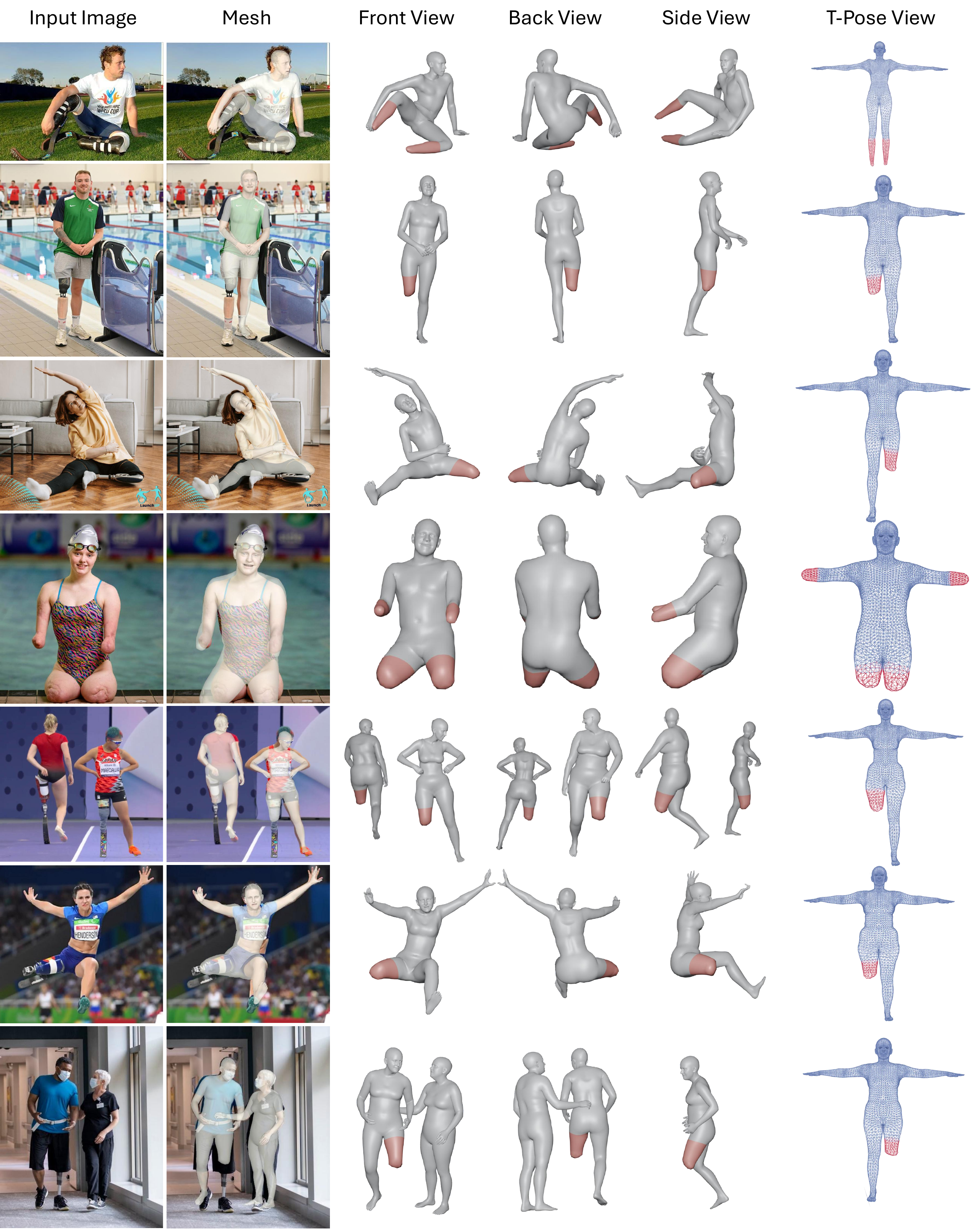}
    \caption{\textbf{Qualitative Evaluation of ResiHMR.} For each input example, we show: (a) the input image, (b) the overlay of SMPL-X in the input view, (c) front view, (d) back view, (e) side view, (f) T-Pose view with model output $ \Theta = \{\boldsymbol{\beta}, \boldsymbol{\psi}, \mathbf{R}, \mathbf{t}, \lambda_r\} $ and $\mathbf{m}_r = \{\, m_k \mid \mathbf{v}_k = (x_k, y_k, z_k, m_k) \in \mathbf{V}_r \,\}$, and (c) (d) (e) and (f) all have residual limb being highlighted in \textcolor{red}{red}.}
    \label{fig:casestudy}
    \vspace{-1.5em}
\end{figure*}

\subsection{Evaluation Protocol}

\noindent\textbf{Baselines.}
We evaluate ResiHMR on the LDPose-LimbLoss Evaluation Dataset with two baselines: SMPLify-X~\cite{smplifyx2019}, an optimization-based SMPL-X fitting method, and HSMR~\cite{hsmr}, a recent regression-based model that uses a biomechanically informed SKEL representation and HMR2.0 training~\cite{HMR20}. Together, these methods cover both optimization and regression paradigms for single-image HMR.
Intact keypoints correspond to the 25 standard OpenPose body keypoints~\cite{openpose}. Residual limbs are referred to 8 residual endpoints defined in~\cite{ldpose}. 
Accordingly, we report three metrics evaluating: (i) intact-body structure (25 keypoints), (ii) the full anatomical keypoint set (33 keypoints), and (iii) residual-limb termination geometry (8 keypoints).

\noindent \textbf{Evaluation metrics.}
We report two metrics, 2D MPJPE and mIoU, to jointly assess structural accuracy and silhouette agreement. 2D MPJPE is the mean per-joint reprojection error between mesh joints and ground-truth 2D keypoints, which we compute separately for intact body joints, residual-limb endpoints, and intact limb joints. mIoU is the mean intersection-over-union between a binary mask rendered from the reconstructed mesh and the manually annotated body mask. Together, these metrics measure alignment between the recovered mesh and the observed anatomy in the image plane and the plausibility of the overall body silhouette for individuals with limb loss.

\subsection{Quantitative Results}

Table~\ref{experiment} reports results on the LDPose-LimbLoss Evaluation Dataset. Existing HMR methods assume intact-limb topology and do not explicitly predict residual-limb endpoints or stump geometry. Therefore, for all non-ResiHMR baselines, residual-limb accuracy is evaluated using a naive midpoint proxy on the corresponding limb segment under a unified protocol.

ResiHMR improves over both backbones. Under the SMPLify-X backbone, it reduces Body Kpts MPJPE from 47.67 px to 41.77 px, Res-Limb MPJPE from 129.59 px to 98.36 px, and Intact Kpts MPJPE from 41.32 px to 37.40 px, while increasing mIoU from 0.662 to 0.703. When combined with HSMR, ResiHMR further reduces Body Kpts MPJPE from 28.27 px to 24.75 px and Res-Limb MPJPE from 73.61 px to 23.19 px, and improves mIoU from 0.705 to 0.741, while maintaining comparable Intact Kpts accuracy (24.56 vs. 24.87 px).\footnote{ResiHMR under HSMR backbone will be detailed in the project page.} Across all methods, ResiHMR (HSMR) achieves the best Body Kpts and Res-Limb results, while CameraHMR attains the highest mIoU (0.752).

These results show that topology-adaptive optimization reduces the distortions caused by forcing intact-limb models to explain missing distal anatomy, leading to better global body alignment. More importantly, ResiHMR is the only method that explicitly models residual-limb endpoints rather than relying on a fixed midpoint proxy, which explains its large advantage on residual-limb localization. The remaining error mainly comes from imperfect initialization and the large image-scale variation in the 255-image dataset, which amplifies pixel-based MPJPE. Overall, ResiHMR substantially improves residual-limb reconstruction while preserving competitive performance on intact anatomy and silhouette quality.

\subsection{Qualitative Results}

Figure~\ref{fig:failedcase} compares ResiHMR with SMPLify-X and HSMR. Because both baselines assume intact limbs, they often hallucinate missing distal segments or collapse the amputated limb into implausible shapes. In HSMR, the full-limb prior can also distort the contralateral leg when a prosthesis or short residual limb is present. By contrast, ResiHMR follows the observed residual-limb endpoints and removes unsupported distal geometry, yielding more stable and anatomically coherent poses.

Figure~\ref{fig:casestudy} further shows that across diverse subjects, poses, and amputation levels, ResiHMR consistently recovers plausible global pose, accurate residual-limb location and length, and smooth watertight stump surfaces from multiple views. These qualitative results match the quantitative findings: unlike existing HMR methods, ResiHMR explicitly models residual limbs and resolves the characteristic failure mode of hallucinated full limbs and distorted contralateral joints.

\begin{table}[t]
\centering
\footnotesize
\caption{\textbf{Comparison of recent HMR methods and ResiHMR under either SMPLify-X or HSMR as backbone.} Best results are shown in bold. For all other HMR methods, as they do not explicitly predict residual-limb endpoints, we define a naive midpoint proxy on the corresponding limb segment to enable consistent scoring under a unified protocol.}
\vspace{-1em}
\setlength{\tabcolsep}{0.25em}{
\begin{tabular}{l|ccc|c}
\toprule
\multirow{2}{*}[-2pt]{Method} & \multicolumn{3}{c|}{2D MPJPE $\downarrow$} & \multirow{2}{*}{mIoU $\uparrow$} \\
\cmidrule(lr){2-4} 
& Body Kpts & Res-Limb & Intact Kpts & \\
\midrule

\makecell[l]{TokenHMR~[CVPR24]~\cite{tokenhmr}}
& 34.79 & 102.34 & 31.73 & 0.717  \\
\makecell[l]{CameraHMR~[3DV25]~\cite{camerahmr}}
& 29.26 & 78.13 & 25.56 & \textbf{0.752}  \\
\makecell[l]{PromptHMR~[CVPR25]~\cite{prompthmr}}
& 51.07 & 102.48 & 46.88 & 0.751  \\

\makecell[l]{HSMR~[CVPR25]~\cite{hsmr}}
& 28.27 & 73.61 & \textbf{24.56} & 0.705  \\
\makecell[l]{SMPLify-X~[CVPR19]~\cite{smplifyx2019}}
& 47.67 & 129.59 & 41.32 & 0.662 \\
\midrule
\makecell[l]{ResiHMR (SMPLify-X)}
& 41.77 & 98.36 & 37.40 & 0.703  \\
\makecell[l]{ResiHMR (HSMR)}
& \textbf{24.75} & \textbf{23.19} & 24.87 & 0.741 \\

\bottomrule
\end{tabular}
}
\label{experiment}
\end{table}

\section{Future Work}
\label{sec:future}

ResiHMR opens two main directions for future work. First, the field still lacks ground-truth 3D datasets for people with limb loss, and calibrated multi-view or marker-based capture would enable stronger supervision and standardized evaluation. Second, learning residual-limb priors directly from data, for example with generative or diffusion models of stump geometry, soft tissue, and limb--socket interfaces, could improve anatomical realism while reducing reliance on optimization. More broadly, we view ResiHMR as a foundation for inclusive human modeling in graphics, vision, rehabilitation, and accessible healthcare.

\section{Conclusion}
\label{sec:conclusion}

We presented ResiHMR, a single-image residual-limb-aware 3D human mesh recovery framework that explicitly modeled residual-limb anatomy instead of relying on fixed intact-limb topology. ResiHMR combined a topology-adaptive Residual Anchor-Factor Optimization component with a Residual-Limb Reconstruction component to produce anatomically coherent meshes and residual-limb-aware 3D body representations that better matched prosthetic biomechanics. On the curated LDPose-LimbLoss Evaluation Dataset, ResiHMR improved intact-joint 2D MPJPE from 41.32 to 37.40 pixels and mask mIoU from 0.662 to 0.703 under the SMPLify-X backbone, and reduced residual-limb 2D MPJPE from 73.61 to 23.19 under the HSMR backbone, while remaining the only method that localized residual-limb endpoints. These results showed the value of amputation-aware kinematics and geometry for inclusive 3D human digitization and indicated new opportunities in prosthetics, rehabilitation, and accessible human-computer interaction.

\section*{Acknowledgement}

This research is funded in part by ARC-Discovery grant (DP220100800 to XY) and ARC-DECRA grant (DE230100477 to XY). We sincerely thank The Advance Queensland Industry Research Projects (AQIRP) for their invaluable support and guidance throughout the development of this work. We thank all anonymous reviewers and ACs for their constructive suggestions.

{
    \small
    \bibliographystyle{ieeenat_fullname}
    \bibliography{main}
}


\clearpage
\setcounter{page}{1}
\maketitlesupplementary

\section{Broader Impact}
\label{sec:BoarderImpact}
Our work aims to advance accessibility and disability inclusion in computer vision by enabling anatomically valid 3D human modeling for individuals with limb loss, an underrepresented population in existing HMR datasets~\cite{h36m, amass, werling2023addbiomechanics, tripathi2023ipman} and methods~\cite{econ, hsmr, HMR20}.

By explicitly reconstructing residual-limb geometry and adapting kinematic structure to limb-loss topology, ResiHMR offers a step toward more equitable and accessible 3D human modeling for applications that extend beyond graphics and pose estimation. The potential positive outcomes extend beyond graphics or pose estimation. Accurate residual-limb geometry is relevant to prosthetic design, rehabilitation assessment, parasport classification, and human-centered AI technologies that must account for bodies with varied physical structures. Our method offers a technical foundation that may support these domains when applied responsibly and in collaboration with clinical or biomechanics experts.

At the same time, modeling real people, particularly individuals with limb loss, requires careful ethical consideration. Residual-limb data may be sensitive, and any downstream clinical or biomechanical use should involve appropriate oversight, domain experts, and consent. Our method does not infer medical diagnoses nor replace clinical assessment; instead, it provides a technical capability that may complement future accessible technologies when used responsibly.

Our work also highlights the broader need for inclusive datasets, evaluation protocols, and modeling assumptions in human-centric AI. We hope it encourages the community to consider disability inclusion as a core component of fair and responsible human modeling, ultimately contributing to AI systems that serve a wider and more diverse range of users.

\section{Importance of Residual-Limb Modeling}
\label{sec:importance}
From the perspective of parasport science, clinical biomechanics, and disability-movement analysis, incorporating the residual limb as an explicit component of the human body model is not merely beneficial, it is essential. The residual limb is an active, load-bearing, and dynamically expressive structure that influences posture, balance control, segment coordination, socket alignment, compensatory strategies, and whole-body movement organization. Ignoring or collapsing this structure, as in methods that hallucinate intact limbs (e.g., HSMR \cite{hsmr}) or truncate geometry to the nearest joint (e.g., AJAHR \cite{ajahr}), fundamentally distorts the underlying anatomy and removes information that is critical for understanding how amputee athletes actually move. Residual-limb length, orientation, and stump-surface geometry directly shape gait kinematics, limb-loading asymmetries, and upper and lower body coordination factors central to parasport classification, performance evaluation, and prosthetic design.

In contrast, ResiHMR reconstructs the residual limb as a first-class anatomical entity, producing continuous termination points and watertight stump surfaces that correspond to the individual’s true morphology. This explicit treatment enables anatomically faithful 3D meshes that retain clinically meaningful geometry rather than simplified joint-level surrogates. As a result, ResiHMR provides not only more accurate reconstructions but also scientifically interpretable models that align with real-world biomechanics. This capability is indispensable for inclusive human-pose research and positions our work as the first step toward anatomically grounded and clinically relevant 3D mesh recovery for individuals with limb loss. 

\section{AJAHR vs. ResiHMR} 
\label{sec:AJAHR}

\begin{figure*}[t]
    \centering
    \includegraphics[width=0.9\linewidth]{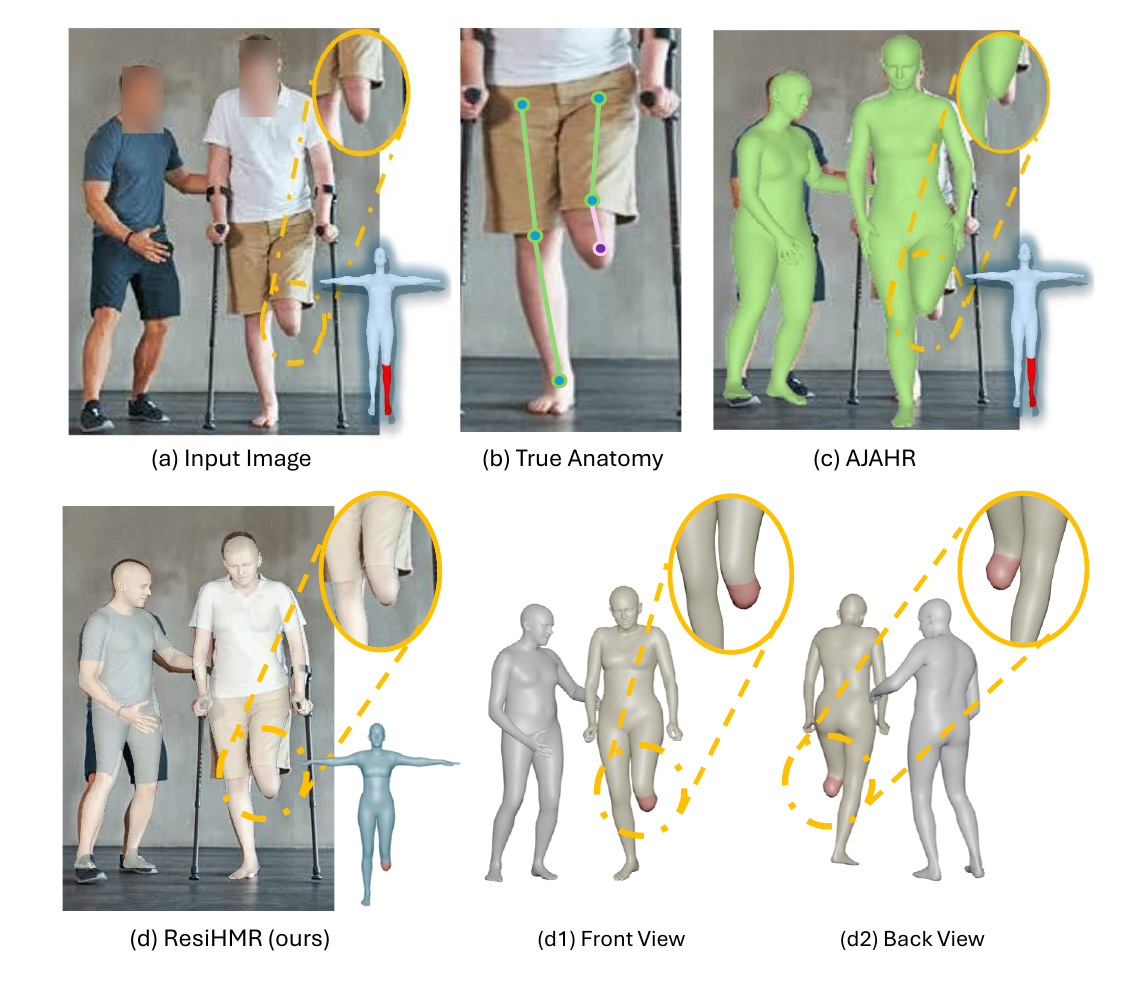}
    \caption{\textbf{A visual comparison of AJAHR and ResiHMR (ours).}  
(a,c) are copied from the AJAHR paper~\cite{ajahr}, where limb loss is represented by collapsing SMPL vertices toward the parent joint, resulting in joint-level truncation.  
(b) shows expert-verified evidence and an additional real image of the same individual revealing a clear below-knee residual limb, indicating that joint-level collapse fails to match the true anatomy.  
(d) shows ResiHMR reconstructions, including a normalized T-pose, with the residual limb highlighted in \textcolor{red}{red} (d1,d2). ResiHMR explicitly estimates the anatomical stump surface rather than collapsing geometry, producing a more realistic and clinically interpretable limb termination.}

    \label{fig:compare}
\end{figure*}

\begin{figure*}[t]
    \centering
    \includegraphics[width=0.95\linewidth]{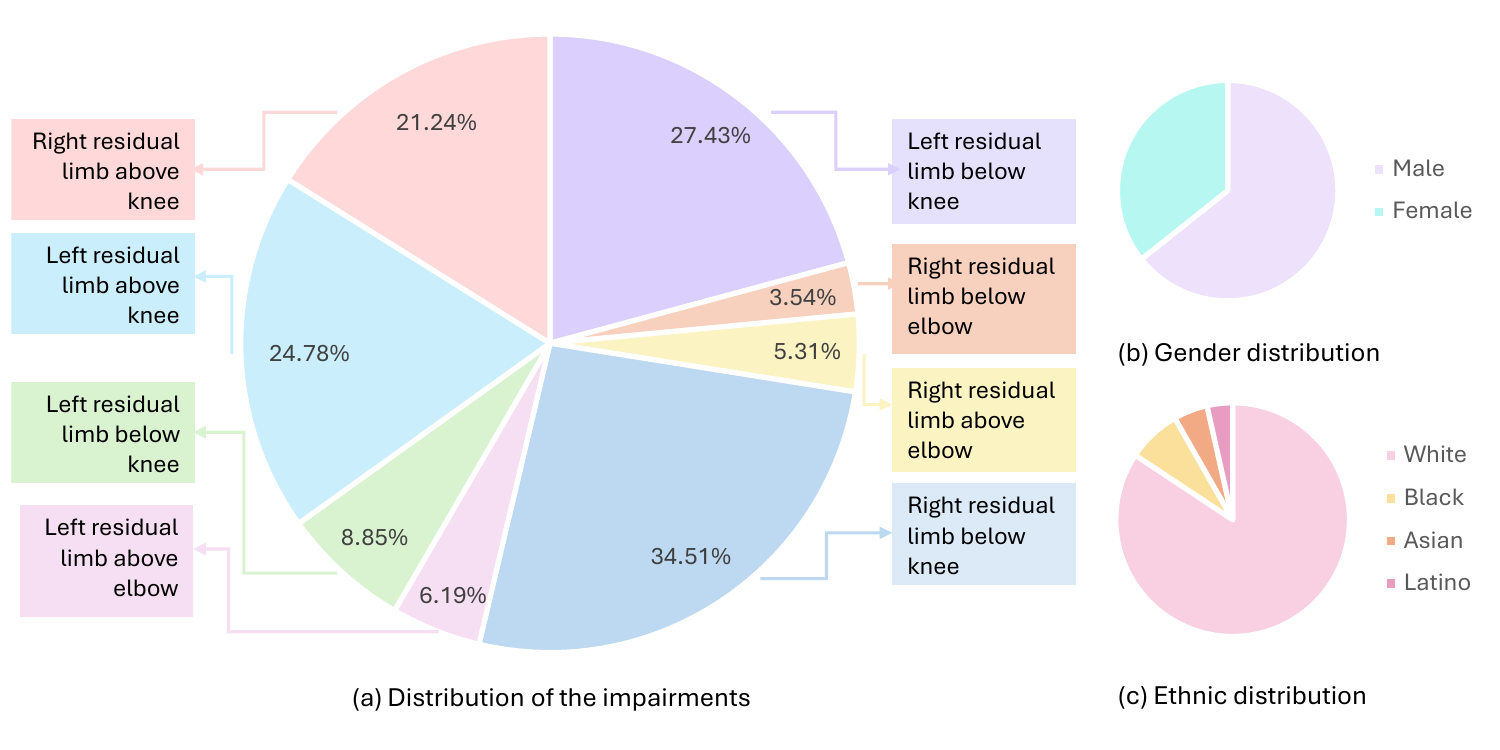}

    \caption{\textbf{Key Statistics of our LDPose-LimbLoss Evaluation Dataset.}  (a) Distribution of residual-limb types, covering upper- and lower-limb amputations across both sides of the body.
(b) Gender distribution.
(c) Ethnic composition of the participants.
Together, these statistics demonstrate the dataset’s demographic and impairment diversity, providing a representative foundation for benchmarking residual-limb–aware 2D/3D human pose and mesh reconstruction.
.}

    \label{fig:dataset_stats}
\end{figure*}

Amputated Joint Aware HMR (AJAHR)~\cite{ajahr} introduces BPAC-Net, which classifies amputated body regions from images and 2D keypoints, and an AJAHR-Tokenizer, a VQ-VAE–style pose tokenizer trained on large-scale intact-body pose datasets together with a synthetic amputee dataset (A3D). In A3D, amputations are generated by modifying SMPL pose parameters: the pose of an amputated joint and all its descendants is zeroed, and the SMPL skinning function collapses distal vertices toward the parent joint. This representation preserves the original SMPL kinematic tree and restricts amputations to occur strictly at existing joint locations. Consequently, AJAHR cannot localize anatomical residual-limb endpoints and produces no stump surface. Instead, the distal mesh simply converges to the nearest intact joint. The supervision in AJAHR, including its amputee training data, therefore reflects only joint-level amputations and lacks ground-truth residual-limb geometry. As a result, while AJAHR can recognize which joint is amputated, the recovered mesh still terminates at the SMPL joint and cannot represent the true residual-limb length, shape, or surface continuity. 
These limitations constrain AJAHR’s applicability to downstream tasks that require explicit residual-limb geometry, such as prosthetic alignment, stump-volume estimation, residuum–socket interaction analysis, or biomechanical evaluation, where continuous stump surfaces and accurate termination points are required.

\subsection{Qualitative Comparison} 
Up to the date of submission, the official AJAHR implementation had not been released, which prevents us from performing a controlled quantitative comparison. For this reason, AJAHR is not included in our experimental evaluation. Instead, we provide a qualitative examination by running ResiHMR on the same input image used in the AJAHR paper and comparing the outputs. As shown in Figure \ref{fig:compare}, the AJAHR results reproduced from the original paper exhibit a joint-level collapse pattern: the predicted mesh contains no residual limb below the left knee, and the left knee itself is collapsed to the proximal joint. To verify the true anatomy of the target individual, we consulted a disability and human-movement expert and also located publicly available images of the same person performing a different action. These independent sources clearly confirm that the subject does have a residual limb below the left knee. Therefore, we include panel (b) in Figure \ref{fig:compare} to illustrate the correct anatomy, which AJAHR fails to represent.
In contrast, ResiHMR reconstructs an anatomically coherent below-knee residual limb, producing a realistic, watertight stump surface as shown in Figure~\ref{fig:compare}(d), with the reconstructed residual limb highlighted in red in (d1) and (d2). This side-by-side comparison underscores the fundamental difference between AJAHR’s joint-level truncation and our explicit residual-limb estimation, demonstrating the necessity of anatomically grounded modeling for amputee subjects.

\subsection{Conceptual Differences}
ResiHMR is designed from a fundamentally different perspective and addresses limitations that AJAHR does not resolve.

\noindent\textbf{Joint-level vs.\ residual-limb level modeling.}  
   AJAHR is \emph{amputated-joint aware}: it predicts whether specific SMPL joints (and their descendants) are amputated and zeroes their rotations, leaving the limb to terminate at the nearest intact SMPL joint.~\cite{ajahr}  
   ResiHMR is \emph{residual-limb aware}: it explicitly models residual-limb endpoints in 2D (via LDPose \cite{ldpose} keypoints) and lifts them into 3D as continuous points along the kinematic chain, rather than snapping them to the nearest SMPL joint. 
   This allows residual limbs to end between canonical joints (e.g., mid-thigh, mid-shank), which is typical for real-world amputations.

\noindent\textbf{Topology collapse vs.\ explicit stump geometry.}  
   In AJAHR, the SMPL topology is never altered: the skeleton and mesh connectivity remain intact, and amputations are simulated by collapsing vertices toward the parent joint.
   This produces a visually amputated region but not a resolved stump surface anatomically.  
   ResiHMR, by contrast, performs a \emph{geometry-based limb termination}: it removes distal vertices, identifies a local boundary ring near the optimized residual endpoint, and reconstructs a smooth, convex, watertight stump surface. 
   As a result, the residual limb becomes an actual geometric structure in the mesh, rather than an implicit collapse of vertices.

\noindent\textbf{Learning regression vs.\ optimization objective.}  
   AJAHR is a learned, token-based regression model, which uses a VQ-VAE tokenizer and a transformer-based HMR architecture trained end-to-end on large-scale synthetic and real data.~\cite{ajahr} 
   Its amputation representation is baked into the SMPL pose parameters and the BPAC-Net classifier.  
   ResiHMR is an optimization-based framework built on SMPLify-X, augmented with a Residual Anchor–Factor Optimization (RAFO) stage. 
   RAFO explicitly optimizes the 3D anchor joint locations and continuous residual-length factors under residual 2D keypoint supervision and anthropometric length priors, enabling anatomically meaningful residual-limb endpoints even without 3D stump ground truth.

\noindent\textbf{Supervision signals.}  
   AJAHR relies on synthetic amputee 3D annotations generated from the SMPL joint-zeroing pipeline; its supervision is therefore aligned with the intact SMPL kinematic tree and joint positions.~\cite{ajahr}  
   ResiHMR instead leverages \emph{residual-limb endpoints} detector developed from LDPose, then uses RAFO plus geometry-based reconstruction to infer stump geometry consistent with these 2D cues and anthropometric segment ratios.
   This makes ResiHMR more suitable for scenarios where accurate residual-limb localization matters (e.g., residual-length estimation, socket-region analysis), even if only 2D annotations are available.

\subsection{Why Joint-Level Truncation Is Insufficient for Downstream Biomechanics}  
Because AJAHR represents amputations through joint-level collapse, implemented by drawing distal vertices toward the parent SMPL joint, the resulting mesh lacks an explicit stump surface and does not provide a measurable residual-limb length along the limb axis. This joint-based truncation preserves the intact SMPL kinematic structure but removes the anatomical information required to characterize the geometry of the residual limb.

As discussed in Section~\ref{sec:importance}, explicit modeling of the residual limb is essential from the standpoint of parasport science, clinical biomechanics, and disability-movement analysis. The residual limb constitutes a load-bearing and dynamically active segment that contributes to balance control, inter-segment coordination, joint kinetics, compensatory strategies, and prosthetic socket alignment. Without a geometric representation of the residual limb, including its termination point, surface continuity, and cross-sectional shape, these biomechanical factors cannot be meaningfully assessed. Consequently, the joint-level representation adopted by AJAHR is unsuitable for downstream applications that rely on physical stump geometry, such as residual-limb volume estimation, residuum–socket interaction modeling, and load-transfer analysis.

In contrast, ResiHMR explicitly estimates residual-limb endpoints and continuous residual-length factors, and reconstructs a watertight stump surface tailored to the individual. This yields anatomically interpretable quantities, including segment length, local cross-sections along the stump, and overall stump geometry, enabling analyses relevant to prosthetic design, rehabilitation, classification research, and broader inclusive biomechanics.

\begin{figure}[t]
    \centering
    \includegraphics[width=1\linewidth]{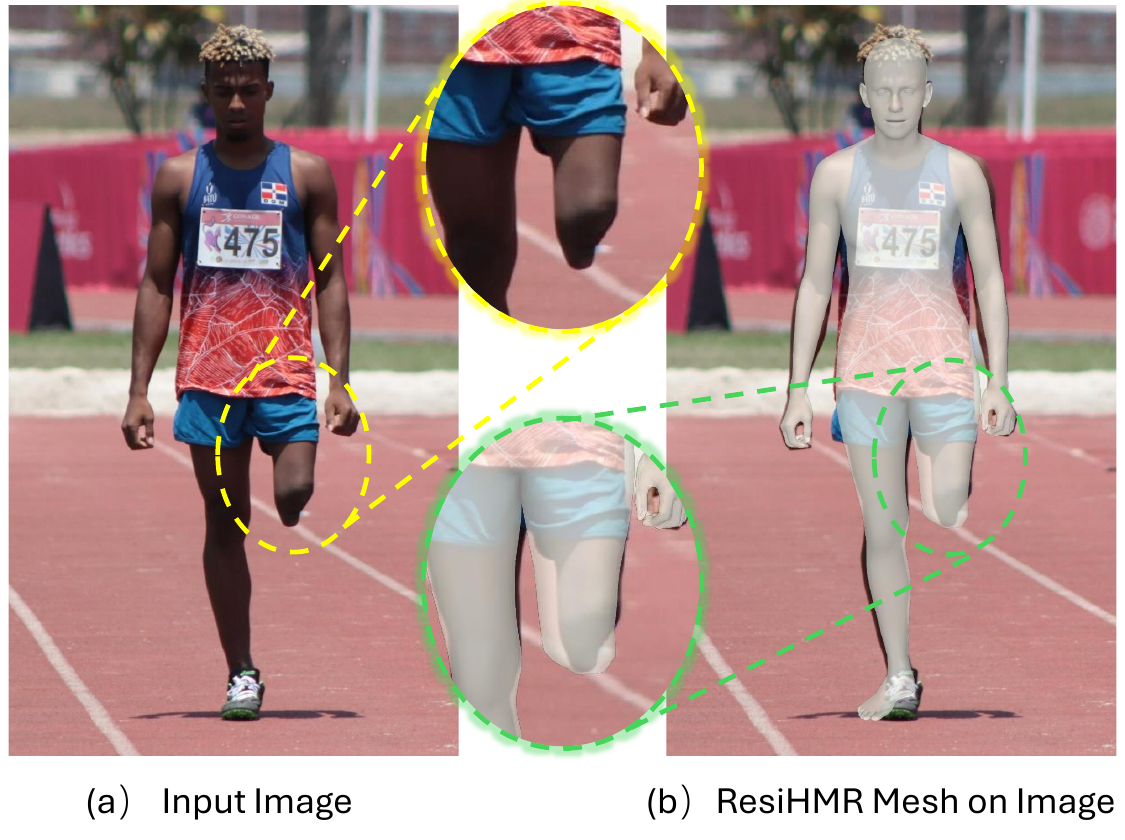}

    \caption{\textbf{Limitation: Residual-limb surface shape approximation.} 
Although ResiHMR accurately localizes the residual-limb endpoint, the reconstructed stump surface adopts a smooth, convex prior that does not fully reflect the subject’s true residual-limb contour (\textcolor{yellow}{yellow}). 
The intact limb is reconstructed faithfully (\textcolor{green}{green}), indicating that the discrepancy arises from limited instance-specific stump shape cues rather than errors in body alignment. 
This limitation reflects the absence of residual-limb 3D training data and highlights a natural growth direction toward modeling individualized residual-limb geometry.}

    \label{fig:limb_shape}
\end{figure}

\begin{figure*}[t]
    \centering
    \includegraphics[width=0.9\linewidth]{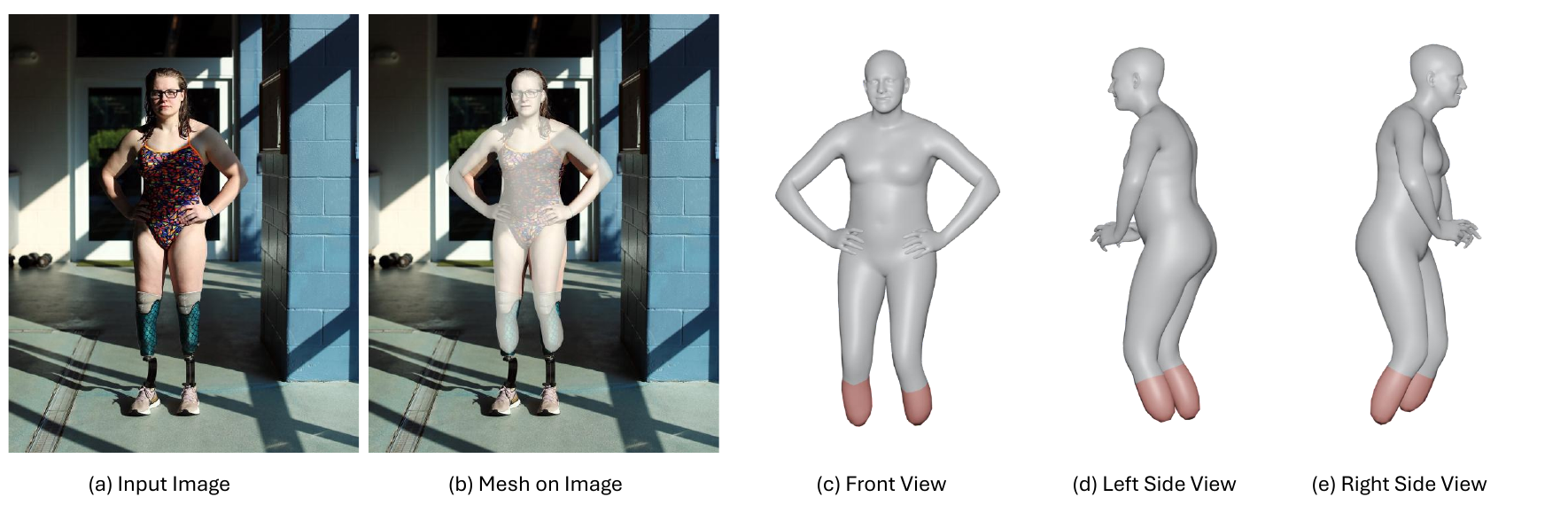}

    \caption{\textbf{Limitation: Depth ambiguity from monocular input.} 
Although the mesh aligns well with the input image in the front view (a–c), the side views (d–e) reveal depth ambiguities inherent to single-image reconstruction. 
Specifically, the arms are positioned in front of the torso rather than resting on the hips, and the upper body exhibits forward flexion that is not visible from the frontal perspective. 
These discrepancies arise from the absence of 3D residual-limb ground truth and the reliance on 2D annotations and healthy-body priors. 
Future 3D datasets with explicit residual-limb geometry may help resolve such ambiguities.}

    \label{fig:depth_ambiguity}
\end{figure*}

\section{ResiHMR with other HMR Methods}
\label{sec:OtherHMRMethods}

\begin{figure*}[t]
    \centering
    \includegraphics[width=\linewidth]{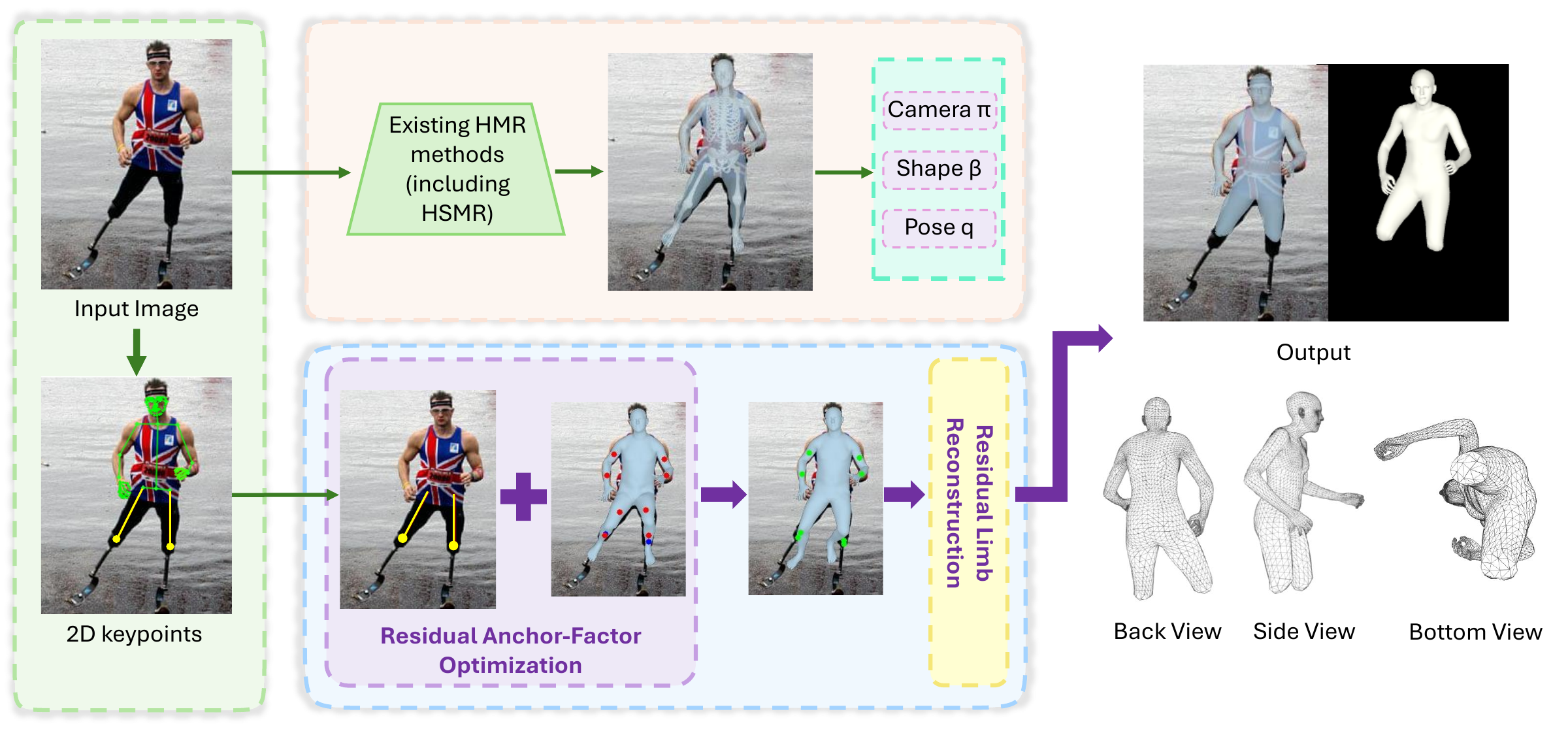}
    \caption{\textbf{Overview of our ResiHMR Framework with other existing HMR methods to initialize the SMPL-X/SMPL}, in  this case, we have HSMR as the exmaple for the regression-based HMR method.}
    \label{fig:pipeline_hsmr}
    \vspace{-1 em}
\end{figure*}

ResiHMR is designed as a backbone-agnostic extension that can be integrated with existing HMR pipelines. 
As illustrated in Figure~\ref{fig:pipeline_hsmr}, we first obtain an initial SMPL-X estimate (camera, pose, and shape) from an off-the-shelf HMR method, which can be either optimization-based (e.g., SMPLify-X) or regression-based (e.g., HSMR). 
This initialization provides a coarse full-body reconstruction under the intact-limb assumption.

Given the input image and corresponding 2D keypoints (including residual-limb endpoints), we then apply our Residual Anchor-Factor Optimization to adapt the kinematic structure by refining anchor joints and residual-limb proportions. 
This step corrects the mismatch between intact-limb priors and amputated anatomy. 
Subsequently, Residual Limb Reconstruction replaces the invalid distal geometry with a smooth, watertight stump surface.

This modular design enables ResiHMR to operate as a plug-in on top of existing HMR systems, extending them from intact-body reconstruction to anatomically coherent modeling of individuals with limb loss. For more details, please visit our project page:

\section{LDPose-LimbLoss Evaluation Dataset Statistics}
\label{sec:statistics}
The LDPose-LimbLoss Evaluation Dataset is designed exclusively for evaluation.  
It is not used for training, pre-training, optimization, hyper-parameter tuning, or any other component of ResiHMR.  
This ensures strict separation between training data and evaluation data, eliminating the possibility of implicit supervision or data leakage.  
All quantitative and qualitative results reported in the main paper are therefore unbiased by this dataset.

We apply rigorous selection criteria to ensure the dataset provides a reliable and fair basis for evaluating amputee mesh reconstruction.  
We include only images in which the majority of the subject’s body is visible, and where residual-limb regions are either clearly observable or can be unambiguously inferred.  
Images with heavy occlusion, severe cropping, or ambiguous limb-termination cues are removed.  
For partially occluded subjects, missing body regions are estimated from nearby visible context to maintain anatomical completeness.  
We also minimize unrelated visual distractors such as cluttered backgrounds, large prosthetic occlusions, or external objects that may interfere with pose estimation.  
These curation steps ensure that evaluation accuracy reflects the algorithm’s performance rather than noise from missing anatomy or extreme occlusion.

The final dataset contains 255 curated images that span a broad variety of residual-limb configurations, amputation levels, residual-limb lengths, activities, genders, and ethnic backgrounds.  
Each image is annotated with 17 standard 2D body keypoints and 8 additional residual-limb endpoints.  
To further facilitate optimization-based mesh fitting, we provide a per-person segmentation mask that isolates the human body region and reduces the influence of clothing, hair, prosthetic devices, equipment, and background objects.  
These annotations create a clean and controlled setting for evaluating residual-limb aware pose and mesh reconstruction models.

All annotations follow the LDPose protocol with extensions specific to limb-loss anatomy.
Annotators were given detailed guidelines on identifying residual-limb endpoints, stump boundaries, and termination cues.  
Cases with uncertain or ambiguous stump locations were independently reviewed by multiple annotators and resolved through consensus.  
When multiple images of the same subject were available, cross-image consistency checks were performed to ensure stable residual-limb endpoint placement.  
This multi-stage quality control procedure ensures that the evaluation labels are precise, consistent, and anatomically meaningful.

The dataset includes demographic and impairment-level diversity, which reduces bias and improves generalizability.  
Finally, precise residual-limb annotations and segmentation masks ensure that reconstructed meshes can be evaluated with respect to anatomically accurate ground-truth cues.

Figure~\ref{fig:dataset_stats} presents the full statistics of the LDPose-LimbLoss Evaluation Dataset. 
The dataset includes 255 images distributed across a wide range of amputation types, genders, and ethnic backgrounds:

\noindent\textbf{Residual-limb types (Figure ~\ref{fig:dataset_stats} (a)).}
    The dataset spans upper- and lower-limb amputations, covering above- and below-elbow and above- and below-knee cases on both sides of the body.
    This diversity supports evaluation across a broad set of real-world amputation configurations.
    
\noindent\textbf{Gender distribution (Figure ~\ref{fig:dataset_stats} (b)).}
    Both male and female subjects are included, enabling analyses that account for gender-related anthropometric variation.
    
\noindent\textbf{Ethnic composition (Figure ~\ref{fig:dataset_stats} (c)).} 
    Participants represent multiple ethnic backgrounds, providing variation in body shape, skin tone, and appearance critical for inclusive pose and mesh reconstruction.

Together, these statistics highlight the \emph{demographic} and \emph{impairment-level diversity} of the dataset, establishing a representative benchmark for residual-limb aware 2D/3D human pose and mesh reconstruction.

\section{Discussion}
\label{sec:discussion}
Although ResiHMR advances anatomically grounded 3D modeling for individuals with limb loss, several limitations remain that reflect the current state of available data and the inherent ambiguity of single-image reconstruction rather than deficiencies of the method.

First, while ResiHMR can reliably estimate continuous residual-limb endpoints for the majority of typical amputation cases, particularly when the residual and intact limbs share broadly similar geometric structure, it cannot yet capture complex limb deformities or highly irregular residual-limb shapes. As illustrated in Figure~\ref{fig:limb_shape}, the reconstructed mesh aligns well with the intact limb but does not fully reproduce the subject’s true residual-limb contour. This limitation arises because the model currently infers stump geometry from smooth convex priors rather than detailed, instance-specific surface cues. We view this not as a shortcoming of the framework, but as a natural growth direction: future work incorporating richer shape priors or multi-view information may enable individualized residual-limb surface modeling.

Second, because there is currently no 3D dataset containing ground-truth residual-limb geometry, ResiHMR relies on 2D annotations and anthropometric priors derived from healthy-body datasets. The lack of 3D supervision introduces depth ambiguity that is inherent to single-image reconstruction. As shown in Figure~\ref{fig:depth_ambiguity}, the mesh appears well aligned in the front view, yet the side view reveals discrepancies: the arm is positioned in front of the torso rather than resting on the hip, and the upper body exhibits forward flexion not visible in the frontal image. Such ambiguities reflect limitations of monocular input rather than the optimization framework. The emergence of future 3D datasets containing residual-limb ground truth would likely mitigate these depth-related deviations and enable more accurate modeling of amputee-specific body postures.

These limitations point to several promising future directions, such as
integrating multi-view information, incorporating uncertainty-aware
optimization, or learning limb-loss–specific 3D priors, while not
detracting from the core contribution of ResiHMR, enabling explicit and
anatomically grounded modeling for individuals with limb loss.

\section{Qualitative Comparison}
\label{sec:quacomp}
\begin{figure*}[t]
    \centering
    \includegraphics[width=\linewidth]{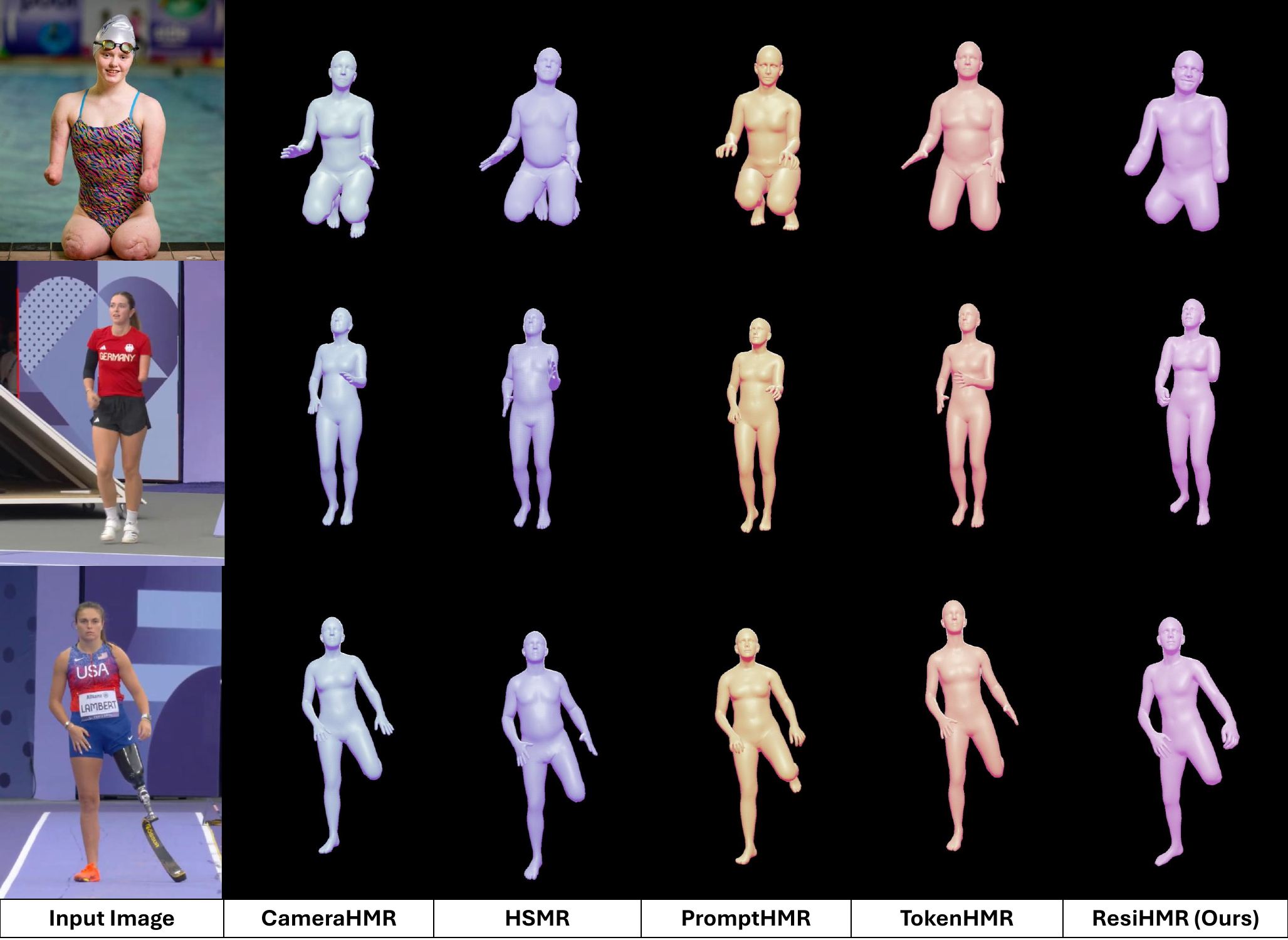}
    \caption{\textbf{Qualitative Comparison of ResiHMR  with SOTA HMR methods.} Please see this in GIF format in project page \href{https://akitaraphael.github.io/ResiHMR/}{\faGithub~\textcolor{purple}{ResiHMR}.}}
    \label{fig:compare_vis}
    \vspace{-1 em}
\end{figure*}

In this section, we qualitatively compare ResiHMR with state-of-the-art HMR methods on individuals with limb loss from single-image inputs.
Representative examples are shown in Figure~\ref{fig:compare_vis}, highlighting the advantages of our approach.

\section{ResiHMR Demonstration}
\label{sec:demo}
\begin{figure*}[t]
    \centering
    \includegraphics[width=\linewidth]{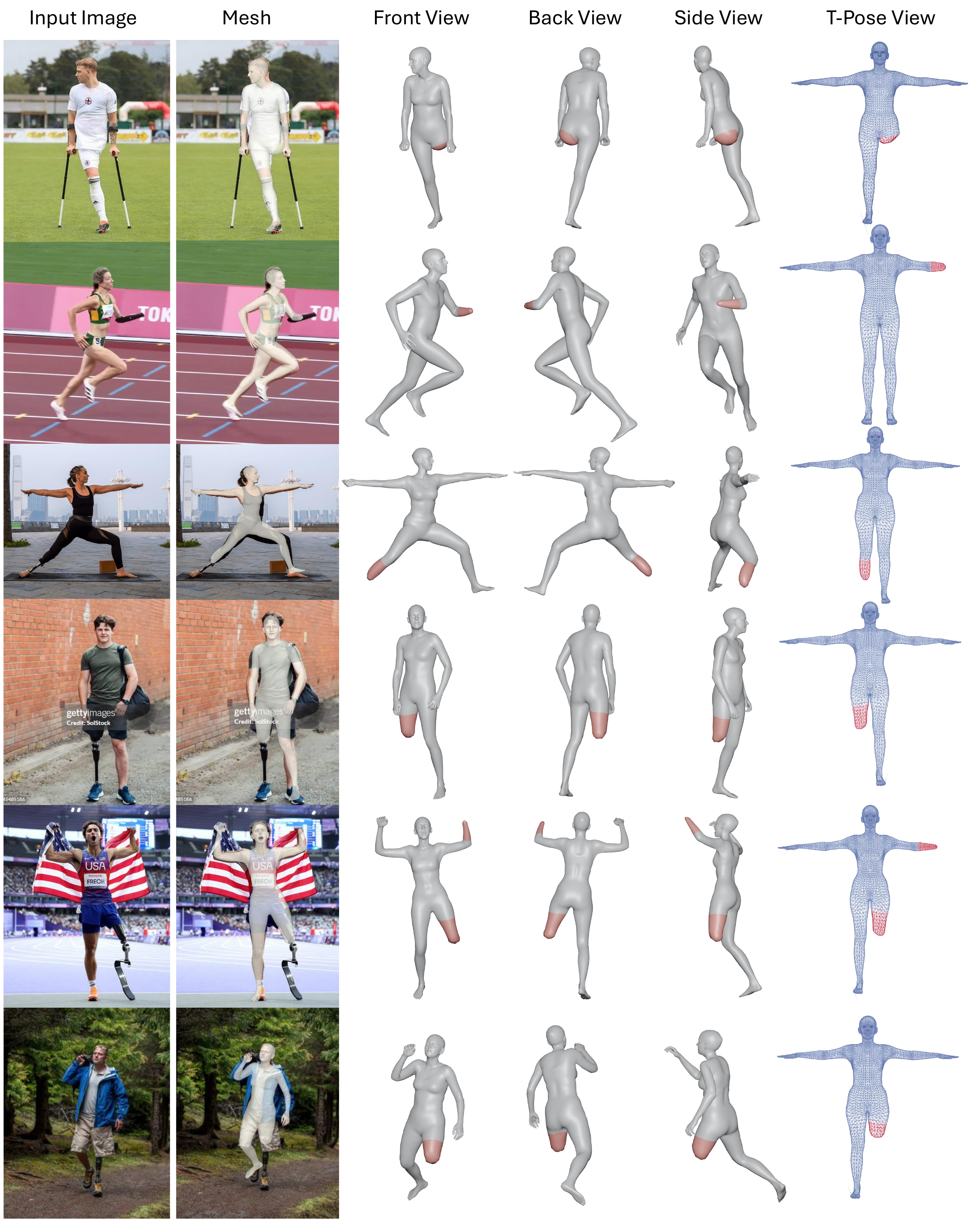}
    \caption{\textbf{More Qualitative Evaluation of ResiHMR.} For each input example, we show: (a) the input image, (b) the overlay of SMPL-X in the input view, (c) front view, (d) back view, (e) side view, (f) T-Pose view with model output $ \Theta = \{\boldsymbol{\beta}, \boldsymbol{\psi}, \mathbf{R}, \mathbf{t}, \lambda_r\} $ and $\mathbf{m}_r = \{\, m_k \mid \mathbf{v}_k = (x_k, y_k, z_k, m_k) \in \mathbf{V}_r \,\}$, and (c) (d) (e) and (f) all have residual limb being highlighted in \textcolor{red}{red}.}
    \label{fig:casevis_append}
    \vspace{-1 em}
\end{figure*}

\begin{figure*}[t]
    \centering
    \includegraphics[width=\linewidth]{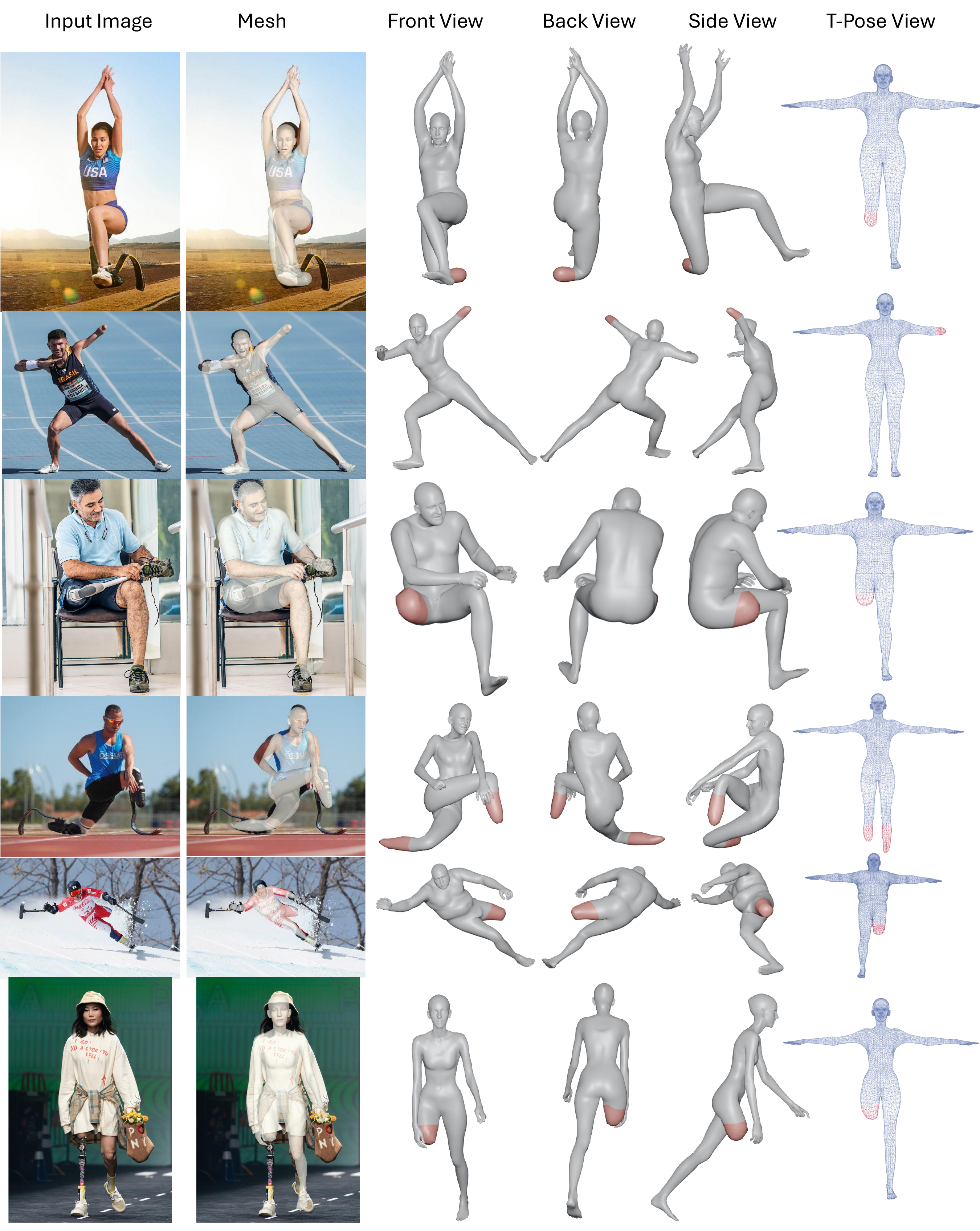}
    \caption{\textbf{More Qualitative Evaluation of ResiHMR.} For each input example, we show: (a) the input image, (b) the overlay of SMPL-X in the input view, (c) front view, (d) back view, (e) side view, (f) T-Pose view with model output $ \Theta = \{\boldsymbol{\beta}, \boldsymbol{\psi}, \mathbf{R}, \mathbf{t}, \lambda_r\} $ and $\mathbf{m}_r = \{\, m_k \mid \mathbf{v}_k = (x_k, y_k, z_k, m_k) \in \mathbf{V}_r \,\}$, and (c) (d) (e) and (f) all have residual limb being highlighted in \textcolor{red}{red}.}
    \label{fig:casevis2}
    \vspace{-1 em}
\end{figure*}

In this section, we present additional qualitative results of ResiHMR to further illustrate its behavior across diverse limb-loss configurations, activities, and viewing conditions. 
Representative examples are shown in Figure~\ref{fig:casevis_append} and Figure~\ref{fig:casevis2}, demonstrating the robustness of our residual-limb modeling pipeline and the anatomical coherence of the reconstructed meshes.

\end{document}